\documentclass[letterpaper, 10 pt, conference]{ieeeconf}  

\IEEEoverridecommandlockouts                              

\usepackage{graphicx} 
\usepackage[dvipsnames]{xcolor}
\usepackage[bb=boondox]{mathalfa}

\title{\LARGE \bf
End-to-end Autonomous Driving using Deep Learning: A Systematic Review}

\author{Apoorv Singh$^{1}$ 
\thanks{$^{1}$Apoorv Singh is from Robotics Institute, Carnegie Mellon University (CMU).
{\tt\small apoorv93singh@gmail.com}}%
}
\begin{document}
\maketitle
\thispagestyle{empty}
\pagestyle{empty}

\begin{abstract}
End-to-end autonomous driving is a fully differentiable machine learning system that takes raw sensor input data and other metadata as prior information and directly outputs the ego vehicle's control signals or planned trajectories. This paper attempts to systematically review all recent Machine Learning-based techniques to perform this end-to-end task, including, but not limited to, object detection, semantic scene understanding, object tracking, trajectory predictions, trajectory planning, vehicle control, social behavior, and communications. This paper focuses on recent fully differentiable end-to-end reinforcement learning and deep learning-based techniques. Our paper also builds taxonomies of the significant approaches by sub-grouping them and showcasing their research trends. Finally, this survey highlights the open challenges and points out possible future directions to enlighten further research on the topic. 

\end{abstract}

\section{Introduction}
Unlike modular autonomous driving stacks, end-to-end autonomous driving benefits from joint feature optimization with minimal human-designed heuristics. These modular stacks, viz., perception, prediction, tracking, planning, controls, et al., are designed to optimize that particular fine-grained task, which may or may not lead to optimizing the primary goal of end-to-end autonomous driving. End-to-end independent driving system research has recently flourished because large-scale datasets are available with minimal data-annotation cost. To train an ML-based model, all we need is a calibrated sensor suite on the car that monitors the control movement of the vehicle by the human driver. This system can theoretically be enabled on any L2/L3 level car already in production. \\

\subsection{History}
\begin{figure}[ht]
\vskip 0.2in
\begin{center}
\centerline{\includegraphics[width=7cm]{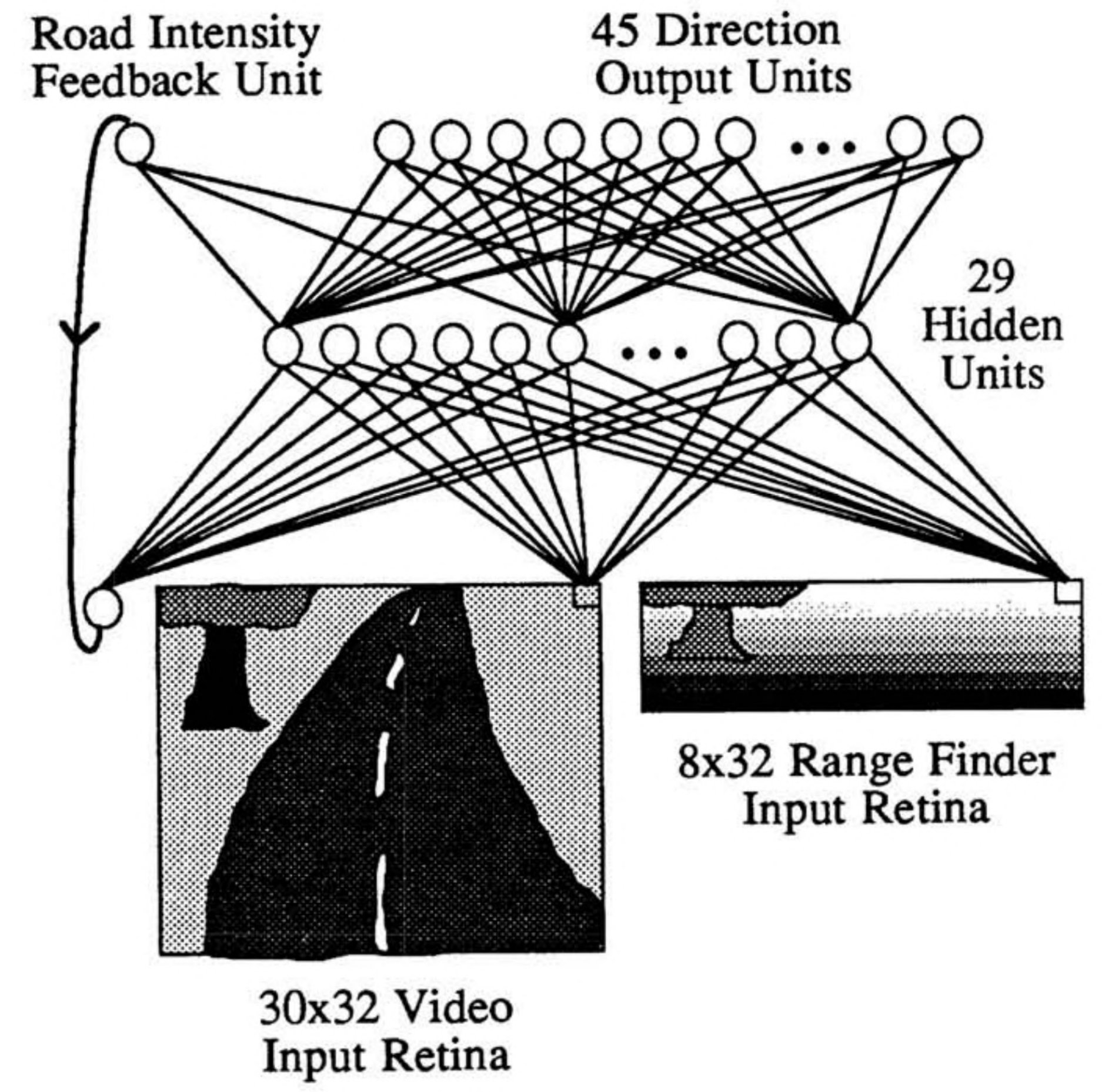}}
\caption{Imitation learning-based end-to-end model architecture from ALVINN \cite{alvin} that takes in perception sensor and outputs control signals.}
\label{fig:alvin}
\end{center}
\vskip -0.2in
\end{figure}

Self-driving cars may seem like a very recent technological phenomenon, but researchers and engineers have built vehicles that can drive themselves for over three decades. Research on computer-controlled vehicles began at Carnegie Mellon in 1984, and the production of the first vehicle, Navlab 1, was formed in 1986. ALVINN, which stands for Autonomous Land Vehicle In a Neural Network. This version of the autonomous car was an end-to-end system where you feed the model a camera image frame (in this case, a series of frames from a trip from Mountain View to Half Moon Bay, a route it’s never seen before) and laser point cloud. The model outputs a value it believes would steer the car correctly on its way. The input layer is divided into three units: two \emph{retinas} and a single intensity feedback unit. The two input channels are camera-based video inputs and LiDAR-based range information. The activation level of each unit in this retina is proportional to the intensity in the blue color band (out of the 3 RGB channels) because it provides the highest contrast between the road and non-road. The second retina consists of $8*32$, which represents the LiDAR point cloud. The road intensity feedback unit indicates whether the road is lighter or darker than the non-road in the previous image. Each of the $1217$ inputs is fully connected to the 29 units in the hidden layer, which are subsequently connected to the $45$ output control signals as shown in Fig. \ref{fig:alvin}.  

In 2016, Nvidia came up with another novel and more complex version of their end-to-end network \cite{e2e_nvidia}. Their approach scaled imitation learning to AlexNet-style deep learning. It demonstrated lane-following across diverse road appearances, sun, rain, and snow. However, this approach was only for steering control and not speed control. Later, in 2018, a company called Wayve came up with a primarily Reinforcement learning-based approach \cite{e2e_wayve, e2d_wayve_2}. They showed they could learn a model with just $10$ episodes of video logs to perform lane-following tasks. Soon, this field of work went into a fast-track mode, and in 2019, Intel came up with an approach to develop a condition-learned policy on navigation commands \cite{e2e_intel, e2e_v1} to navigate Manhattan-style intersections. Their system was evaluated on simulators and an RC car. Later, in 2019, a paper from ETH Zurich \cite{e2e_eth} developed a system that can operate on Manhattan turns and more generalized autonomy on an open loop system. Uber ATG has also developed a system \cite{e2e_atg} that used LiDAR and HD maps and input and outputs motion plan, which was evaluated on an open loop system. Here, modular autonomy tasks were learned as an auxiliary tasks to guide the end-to-end learning. 
 
\begin{figure}[ht]
\vskip 0.2in
\begin{center}
\centerline{\includegraphics[width=\columnwidth]{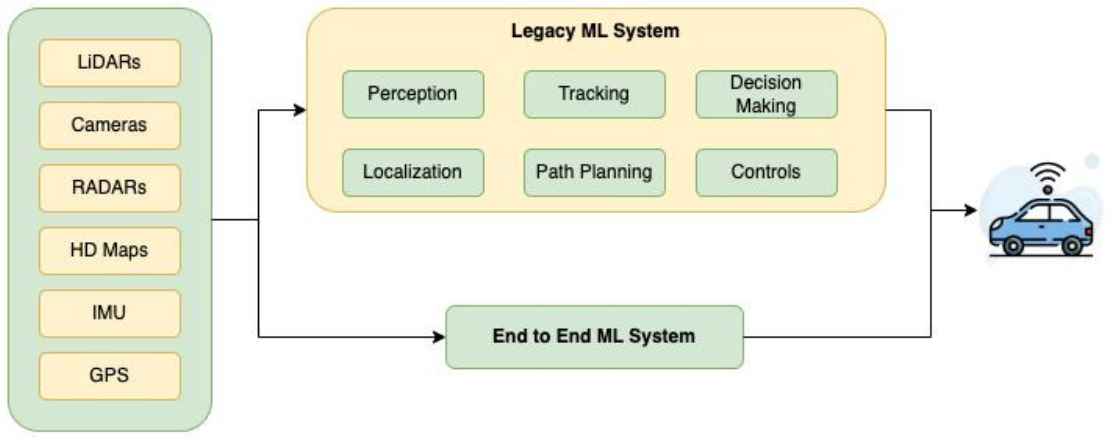}}
\caption{Comparison of Modular Autonomous driving tasks vs single-model of end-to-end autonomous driving task.}
\label{fig:e2e_arch}
\end{center}
\vskip -0.2in
\end{figure}

\subsection{Contributions}

This paper has three key contributions:
\begin{itemize}
    \item We provide comprehensive background knowledge for understanding end-to-end autonomous driving approaches. We started by summarizing the first technique ever applied in autonomous driving tasks. Moreover, we summarized the competing alternate approaches of the modular autonomous driving problem solutions. We also cover evaluations and benchmarks used for end-to-end autonomous driving tasks.
    \item Next, we extensively covered end-to-end approaches by categorizing them in four sections of \emph{Imitation Learning}, \emph{Reinforcement Learning}, \emph{End to end Autonomous driving with the auxiliary task}, and \emph{Teacher Student Paradigm}. Moreover, before diving into detail, we covered a paragraph on theoretical knowledge for the readers with mathematical modeling. 
    \item We discussed key challenges for all the methodologies and approaches covered to enlighten readers with prospective future research ideas. In addition, we discussed a few of the open-ended research problems that we foresee would be immediate next steps in end-to-end autonomous driving research.
\end{itemize}
A few surveys have addressed open end-to-end autonomous driving \cite{survey_1, survey_2, survey_3, survey_4}, but none cover significant recent advances in the fields. Moreover, they did not explore many of the research directions we propose in this paper. Some of the prior research papers only covered specific topics within the end-to-end system, but we emphasized covering all the major technologies that are active right now while ignoring/ partially covering dated techniques to keep readers interested.  

\subsection{Problem Formulations}
In straightforward words, with end-to-end autonomous driving, we are trying to learn a function that maps Input State $X$, of magnitude $\sim 10^8$, to the output state $Y$, which has a combination of $\sim 10^1$ states. Where input data, $X$, may include different combinations of multiple perception sensors, viz., cameras/ RADARs and LiDARs, it may also include metadata like HD maps, vehicle state, sensor calibration information, GPS, IMU data, etc. However, the output state, $Y$, combines control signals like brake/ accelerate/ steering and other cues like indicators/ headlights. \\
Carla \cite{carla} has a simplified version of this where an agent interacts with the environment over discrete time steps. At each time step, the agent gets an observation $O_t$ and must produce an action $A_t$. The observation, $O_t$ is a tuple of sensory inputs, and the action, $A_t$, is a three-dimensional vector representing the steering, throttle, and brake. 

\subsubsection{Input Sensor Modalities}
Primary perception sensor modalities are Cameras, LiDARs, and RADARs. Generally, an autonomous vehicle uses different combinations and quantities of these sensors, depending on their SAE Autonomy level, to perceive its environment. Generally, up to level 3 SAE level cars rely more on the Camera-RADAR system because of their value for money in semantic and depth information captured. However, the L3+ SAE level of autonomous cars depends more on the LiDAR-Camera system since they are building absolutely safe autonomous cars with very high recall to be driven without human drivers. LiDAR's much denser and semantically more informative point cloud data, compared to the RADAR, comes at a higher upfront cost of the sensor, which can only be budgeted for the L3+ systems. In addition to these sensors, autonomous cars use metadata that performs transformation from one sensor to another. These calibrated transformations can get all the sensor information in a single unified ego plane to perform the perception task.  \\
In addition to the series of perception sensors, L3+ autonomous cars heavily rely on HD Maps, which hold centimeter-level accuracy of the roads they drive on. In addition to the road boundaries, these maps may include other semantic information like crosswalks, lane markings, intersections, traffic-light positions, etc. However, due to the geographic scales of L2-L3 cars, generating HD maps can get too expensive. They mostly compute such local maps online using perception sensors. GPS and IMU sensors are used to geo-localize autonomous cars so that they can locate themselves within the predefined map and proceed with their further downstream tasks of trajectory prediction and trajectory planning. \\
Recently, more diverse technology of sensors are being utilized for autonomous cars, for example, fish-eye cameras, which can detect close-range objects very well; velocity-LiDARs: These are LiDAR-like sensors that give dense point-cloud as any other LiDAR with the added benefit of velocity attribute per point, just like RADAR, using Doppler effect. Certain indoor autonomous robots also have stereo cameras, which can generate short-range dense point clouds. \\
\subsubsection{Output Sensor Modalities}
Output modality is fairly simple; it outputs control signals like steering angle, acceleration/ braking information, and meta signals like turn lights, headlights, etc. These annotations can be easily taken from any human-driven car, thus making it an ideal problem formulation for Imitation learning. However, \emph{explainability/ interpretability} is one of the major drawbacks to propagating raw sensor data information all the way to this fairly small dimensionality of control signals. When a problem arises with autonomous car behavior with such an end-to-end machine learning system, it is extremely hard to figure out targeted improvements on the software, leaving us just feeding in and learning from more and more data. 
To increase the system's interpretability, some authors have added auxiliary losses to supervised complete end-to-end systems, which we will cover in the later section. Another approach that few of the authors have taken is to disentangle the control stack from this end-to-end system and, instead, output the future waypoints, defined as future locations of the ego vehicle. Then, control operates on these waypoints as a downstream task to execute the action. 

\section{Modular Approach for Autonomous Driving}

A traditional, modular autonomous driving software is a complex system comprising various sub-modules, each responsible for a specific aspect of the driving process. These sub-modules work together to enable a vehicle to navigate, perceive the environment, and make decisions without human intervention. Some of the critical sub-modules of autonomous driving software include: 
\begin{itemize}
    \item Mapping and HD-Mapping: It refers to creating and maintaining detailed, high-definition maps of the road and surrounding environment. Unlike traditional navigation maps used by human drivers, autonomous driving maps are much more detailed and include information crucial for the specific needs of self-driving systems. They include information about lane boundaries, road curvature, traffic signs, traffic signals, lane-change rules, and road gradients. \cite{hdmap} HD maps are created with centimeter-level accuracy to ensure the autonomous vehicle can position itself accurately within its environment in the localization stack downstream.
    \item Perception: It refers to the ability of a self-driving vehicle to understand and interpret its surrounding environment. It includes tasks like sensor fusion, vehicle and pedestrian detection, pixel-level segmentation, and traffic-light/ other road-sign detection. Primarily, it is based on multiple quantities and a combination of camera, LiDAR, and RADAR-based sensors \cite{wod}. 
    \item Localization: It refers to the ability of a self-driving vehicle to accurately determine its position and orientation (pose) within its environment. \cite{localization} Critical components used are the Global Positioning System (GPS), Inertial Measurement Unit (IMU), and perception sensors via map matching, which involves comparing sensor data with the HD map to refine the vehicle's position estimate.
    \item Object Tracking: It refers to the ability of self-driving vehicles to monitor the past movement of objects in their environment. This is crucial to make predictions in the downstream task, for which past states are critical. This stack is essential for occluded road objects as we can still maintain our expectation of the object even though it is not visible in the perception stack. \cite{sort} This stack involves data association and kinematics prediction of the perception outputs. 
    \item Behavior Planning: It refers to the ability of self-driving vehicles to forecast the future actions and movements of various road users, such as pedestrians, cyclists, other cars, and even animals. \cite{prediction} Large-scale models can be generated for this module by good detection and tracking modules in the offline system that can generate trajectory prediction data for free. Autonomous systems account for uncertainty in behavior prediction. They consider potential variations in predicted behaviors, accounting for factors like sensor noise, unpredictable human actions, and changes in traffic conditions.
    \item Path Planning: It refers to the ability of self-driving vehicles to determine a safe and optimal path for a self-driving vehicle to navigate from its current location to a desired destination while avoiding obstacles, adhering to traffic rules, and considering the dynamics of the environment. Based on the current vehicle position, the desired destination, and the map and sensor information, the path planning algorithm generates a preliminary path that connects the vehicle to the goal while avoiding obstacles. The cost function considers distance, comfort, safety, energy efficiency, and collision risk factors. The algorithm searches for a path that minimizes this cost.  The most common approach for planning in modular pipelines involves using sophisticated rule-based designs, often ineffective in addressing the many situations that occur while driving. 
    \item Decision Making: It refers to the ability of self-driving vehicles to select appropriate actions and behaviors to navigate through various traffic scenarios while prioritizing safety, efficiency, and adherence to traffic rules. It works closely with path planning and involves risk assessment, scenario analysis, behavior prediction, traffic rules and regulations, ethical and moral considerations, etc.
    \item Control: It refers to the mechanisms and algorithms that translate high-level decisions and planned trajectories into specific actions taken by the vehicle's actuators to maneuver the vehicle physically. These controls ensure that the vehicle follows the intended path, accelerates, decelerates, and handles various driving scenarios while maintaining stability, safety, and comfort. Controls bridge the gap between decision-making and the physical behavior of the vehicle. It includes sub-components such as actuators, vehicle dynamics, speed control, stability and handling, smooth transitions, human comfort, etc.
    \item Simulation and Testing: It is essential to assess self-driving systems' performance, safety, and functionality before being deployed on real roads.

    These systems and their dependency graph can be seen in the sequential order in the Legacy ML System as shown in Fig. \ref{fig:e2e_arch}
\end{itemize}

\section{Benchmarking/ Evaluations}
As unanimously accepted in software practice, to first develop the testing requirement before working on the software itself, let us look at the evaluation requirement for the end-to-end systems before diving further into the major methods. Evaluations can be divided into two types: 1. Closed-loop evaluations and 2. Open-loop evaluations. Former is defined in an online simulator; however, former is computed offline on human driving datasets. End-to-end systems can be easily evaluated on an online simulator like CARLA \cite{carla}; many modular perception techniques rely on the more expensive offline datasets.  

\subsection{Closed-loop evaluations}
\begin{figure}[ht]
\vskip 0.2in
\begin{center}
\centerline{\includegraphics[width=\columnwidth]{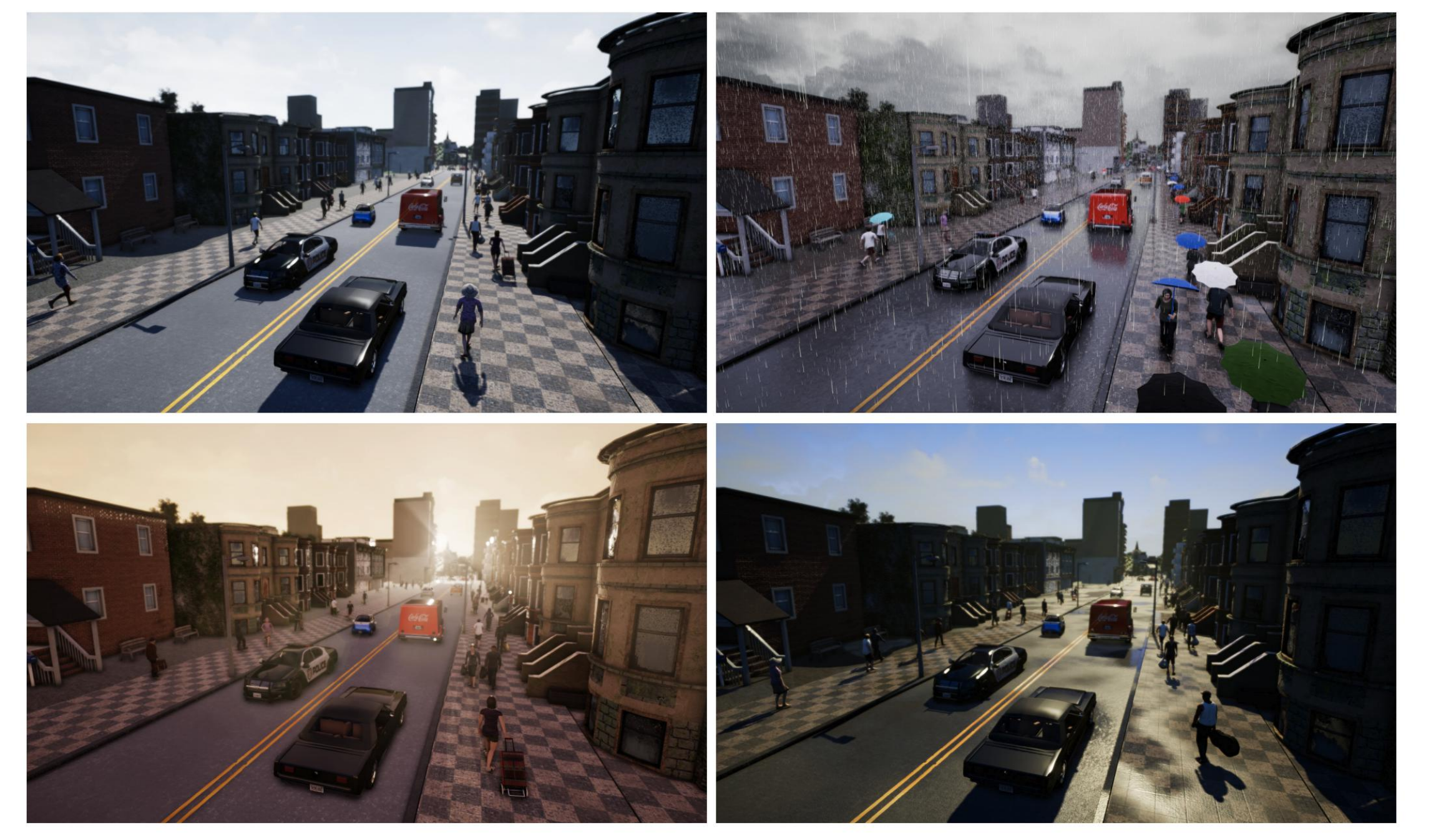}}
\caption{Snapshot from online closed-loop simulator CARLA \cite{carla}.}
\label{fig:carla}
\end{center}
\vskip -0.2in
\end{figure}

To conduct end-to-end tasks is a relatively complicated and dangerous experiment on autonomous cars, at least until we have a reliable and safe software stack for autonomous cars, but to such safe software, we need to train a machine learning model with the correct evaluations. Thanks to online closed-loop simulators \cite{nuplan, carla, torc, e2e_nvidia}, we can test such techniques online while autonomous cars become safer and are widely available for real-world testing. However, even though researchers and industries have tried their best to emulate real-world scenarios as closely as possible in these simulators for constructing towns and driving maneuvers, they still can't be generalized enough for real-world scenes. \\
CARLA \cite{carla} is one of the most widely used free simulators in research environments with diverse traffic and weather conditions, as shown in \ref{fig:carla}. It supports autonomous driving systems' development, training, and detailed performance analysis. They support modular system design (perception/ planning and controls) and end-to-end systems like Imitation learning/ reinforcement learning. For Imitation learning, this dataset includes $D = {O_i, C_i, A_i}$, consisting of tuples, an observation, a human-commands, and an action, respectively. The commands are auxiliary signals the driver provides during data collection to indicate their action, just like turn signals. In this dataset, four commands are used: follow the lane (default),  drive straight at the next intersection, turn left at the next intersection, and turn right at the next intersection. The observations are sensory inputs from multiple sensors. This simulator also injects noise during data collection to increase the robustness of the learned policies. The dataset trains a deep network to predict the expert’s action $A_i$ given an observation $O_i$ and a control command $C_i$. \\

For reinforcement learning-based problems, the vehicle has to reach a goal guided by high-level commands from the topological planner. The episode is terminated when the vehicle reaches the goal when the vehicle collides with an obstacle, or when a time budget is exhausted. The reward is a weighted sum of five terms: positively weighted speed and distance traveled towards the goal, overlap with the opposite lane, negatively weighted collision damage, and overlap with the sidewalk. \\
The tasks are set up as goal-directed navigation: an agent is initialized somewhere in town and has to reach a destination point. In these experiments, the agent can ignore speed limits and traffic lights. The below experiments are conducted in multiple towns, but the towns are mutually exclusive for training and validation datasets. They organize the tasks in order of increasing difficulty as follows:
\begin{itemize}
    \item Straight: Destination is straight ahead of the starting point, and there are no dynamic objects in the environment.
    \item One turn: Destination is one turn away from the starting point; no dynamic objects. 
    \item Navigation: No restriction on the location of the destination point relative to the starting point, no dynamic objects.
    \item Navigation with dynamic obstacles: Same as the previous task, but with dynamic objects (cars and pedestrians).
\end{itemize}
The other most commonly used simulator for closed-loop simulation is nuPlan \cite{nuplan}, which defines its metric as: 
\begin{itemize}
    \item Traffic rule violation measures compliance with common traffic rules. They compute the rate of collisions with other agents, the rate of off-road trajectories, the time gap to lead agents, the time to collision, and the relative velocity while passing agents as a function of the passing distance.
\item Human driving similarity is used to quantify maneuver satisfaction compared to humans, e.g., longitudinal velocity error, longitudinal stop position error, and lateral position error. In addition, the resulting jerk/acceleration is compared to the human-level jerk/acceleration. 
\item Vehicle dynamics quantify rider comfort and feasibility of a trajectory. Rider comfort is measured by jerk, acceleration, steering rate, and vehicle oscillation. Feasibility is measured by violation of predefined limits of the same criteria. 
\item  Agreement between decisions made by a planner and human for crosswalks and unprotected turns (right of way).
\item Goal achievement measures the route progress towards a goal waypoint on the map using L2 distance.
\end{itemize}
The latest trend in generative machine learning has enlightened more realistic simulator generation \cite{diffusion_sim_1, diffusion_sim_2} based on diffusion models; however, they are yet to be mature enough to be literature-wide problems. Some researchers have also developed GAN-based and NERF-based renderings \cite{nerf_sim}.

\subsection{Open-loop evaluations}
Open-loop evaluations have historically been used by modular autonomy tasks independently for waypoint prediction, point-cloud and image segmentation, object detection, etc. Open-loop evaluations refer to testing the systems' performance based on the previously recorded driving instance, which includes the sensor inputs, goal locations, and future trajectories in the form of waypoints. This is not a complete end-to-end system where the evaluations are performed at the end of the perception, prediction, and planning task. Given the sensor inputs and goal locations, the L2 distance is calculated between predicted waypoints compared to the trajectory followed by the real human driver \cite{openloop}. To simplify this problem a bit, researchers from Baidu have decoupled perception and tracking, and they operate on tracker outputs to find the planned trajectory using an MLP network, as shown in Fig. \cite{openloop} 

\begin{figure}[ht]
\vskip 0.2in
\begin{center}
\centerline{\includegraphics[width=\columnwidth]{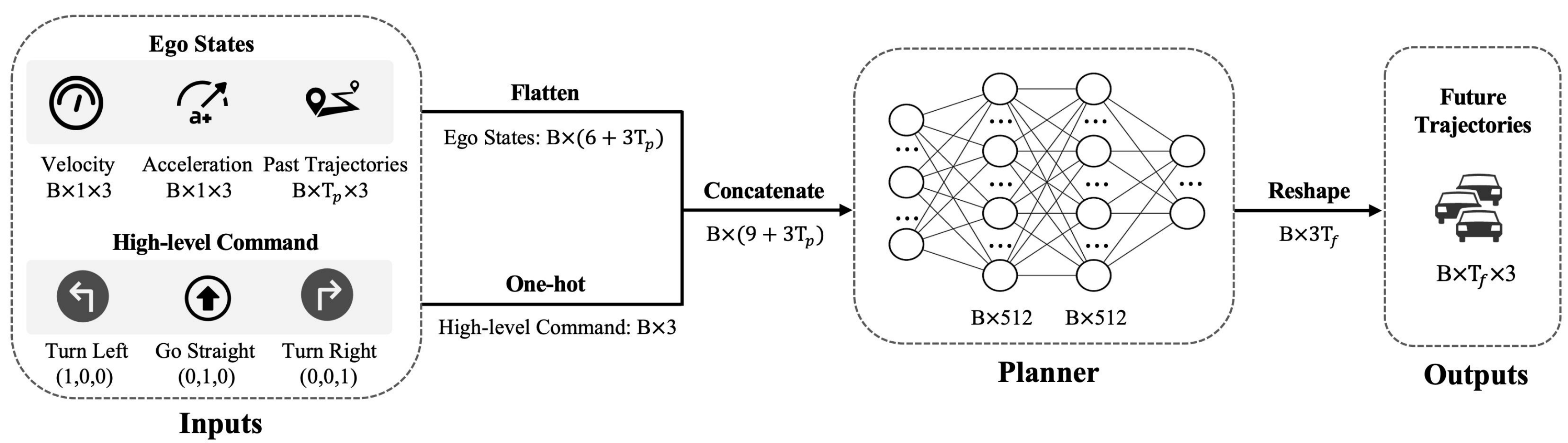}}
\caption{Open-Loop evaluation \cite{openloop} demonstration, where inputs are ego vehicle state, which includes motion trajectories, instantaneous velocities, acceleration, heading angle, etc. Everything that is being outputted from the perception-tracking stack. The output showed planner output in the form of waypoints.}
\label{fig:openloop}
\end{center}
\vskip -0.2in
\end{figure}

\section{Methods}

\subsection{Imitation Learning}
In imitation learning, an agent learns to perform a task by imitating the behavior demonstrated by a human expert or a pre-existing model. Instead of explicitly designing a reward function or defining the optimal policy, the agent learns by observing and mimicking the actions taken by the expert. Imitation learning has several advantages, including the ability to leverage human expertise to solve complex tasks and the potential to provide safer and more interpretable behavior compared to reinforcement learning, where agents learn through trial and error. However, imitation learning also has limitations, such as being sensitive to errors in the expert's demonstrations and not always being able to explore new strategies beyond what the expert has demonstrated. During Training, The agent's objective is to learn a mapping from the input observations to the actions demonstrated by the expert. This is often done using supervised learning techniques, where the agent's model is trained to minimize the discrepancy between its predicted actions and the actions in the expert's dataset. Similarly, during evaluations, the trained agent's performance is evaluated on new, unseen data to determine how well it can imitate the expert's behavior. This evaluation can involve comparing the agent's actions to those of the expert or assessing the agent's success in accomplishing the task. Imitation learning can be further divided into Behavior clone and Inverse reinforcement learning, but first, let's look at its problem formulation.

\subsubsection{Problem Formulations}
In imitation learning, given a dataset $D$ containing a set of trajectories, each trajectory is a sequence of state-action pairs. This action $A$ is taken by the expert at the state $S$ under the guidance of the expert's learned policy $\pi_E$. \cite{survey_4, imitation}The optimization-based strategy of imitation learning is to learn a policy $\pi^*: S \rightarrow A$ that emulates the expert's policy $\pi_E$ by satisfying:
\begin{equation}
    \pi^* = arg min_\pi \mathbb{D}(\pi_E, \pi)
\end{equation}
Where $\mathbb{D}$ represents the similarity between current and expert policies. 

\subsubsection{Behavior Clone}
One of the earliest research dated back to the ALVINN \cite{alvin} uses a fully connected neural network for behavior clone to perform lane following tasks. Nowadays, this fully connected layer is replaced with deep neural networks, but the underlying learning policy remains the same with behavior cloning \cite{e2e_nvidia, imitation_2}. Behavior clone formulates the problem as a supervised learning process to match the learned policy $\pi_\theta$ to the expert policy $\pi_E$:
\begin{equation}
    min_\theta \mathbb{E} || \pi_\theta - \pi_E ||_2
\end{equation}
This is formulated as an L2 loss minimization. As the name suggests, behavior clone-based methods can never explore more scenarios not covered in the supervised dataset; they try to clone the behavior. This leads us to Covariance Shift since it generalizes poorly to the new states due to compounding errors in the action. Covariance shift \cite{covariance} is defined as a specific type of dataset shift often encountered in machine learning. It is when the distribution of input data shifts between the training environment and the live environment. This issue has been mitigated with subsequent papers targeting data augmentation and regularization to add some noise in the action and learning process \cite{imitation_3}. \\

\subsubsection{Inverse Reinforcement Learning (IRL)}
In inverse Reinforcement Learning (IRL), an agent aims to learn the underlying reward function of an environment from observed demonstrations or expert behavior. Unlike standard reinforcement learning, where an agent learns a policy to maximize a reward function, IRL focuses on recovering the reward function by observing how an expert behaves in a particular environment. It's particularly useful when the reward function of an environment is difficult to specify manually but can be inferred from observing human or expert behavior. However, IRL comes with challenges, such as the non-uniqueness of reward functions that explain the same behavior and the sensitivity of the learned reward function to the quality and quantity of expert demonstrations. This is a supervised learning approach where the collected dataset contains trajectories or demonstrations of an expert navigating the environment. Each trajectory consists of the expert's state-action pairs while accomplishing a task. During the training/inference process, features are extracted from the state space to represent important characteristics of the environment. These features are used to represent the states in a more meaningful way and to generalize across different states. IRL aims to find a reward function that explains the observed expert behavior. This is typically formulated as an optimization problem, where the recovered reward function should assign higher values to the states and actions that are consistent with the expert's behavior and lower values to those that are not. Once the reward function is recovered, an optimal policy for the agent can be derived using standard reinforcement learning techniques such as Q-learning or policy gradient methods. The agent maximizes the expected reward according to the recovered reward function. In IRL, the reward function is learned by maximizing the posterior probability of observing expert trajectories:
\begin{equation}
    max_\theta \mathbb{E}_{\pi_E} [G_t|r_\theta] - \mathbb{E}[G_t|r_\theta]
\end{equation}
However, Inverse Reinforcement Learning (IRL) and Imitation Learning comes with its own set of challenges and limitations:
\begin{itemize}
    \item Ambiguity in Expert Behavior: IRL assumes that the observed expert behavior is optimal or near-optimal with respect to some underlying reward function. However, real-world expert behavior can be noisy, suboptimal, or influenced by factors not captured by the reward function, leading to ambiguity in the inferred reward function.
    \item Non-Uniqueness of Solutions: Multiple reward functions can explain the same set of observed expert behavior. This non-uniqueness can make determining which reward function is the "correct" one challenging, leading to uncertainty in the learned reward function and subsequent policy.
    \item Limited Expert Demonstrations: IRL heavily relies on the quality and quantity of expert demonstrations. If the dataset is small, biased, or does not cover a wide range of scenarios, the learned reward function may not accurately represent the true underlying reward structure.
    \item Curse of Dimensionality: As the state space becomes larger and more complex, the ability to accurately recover a reward function from limited expert data becomes increasingly difficult due to the curse of dimensionality. High-dimensional state spaces require large amounts of data to capture the intricacies of the reward function.
    \item Computational Complexity: Many IRL algorithms involve optimization problems that can be computationally expensive, especially when dealing with large state spaces or complex reward functions. 
    \item Lack of Exploration: IRL focuses on learning from expert demonstrations and does not inherently provide mechanisms for exploring novel strategies beyond what the expert has demonstrated. This can limit the agent's ability to discover more effective or creative solutions.
    \item Transfer to New Environments: Learned reward functions are often specific to the environment in which they were inferred. Transferring a learned reward function to a new, unseen environment can be challenging due to differences in dynamics, state space, and other factors.
\end{itemize}

\subsection{Reinforcement Learning}
In reinforcement learning, an agent learns how to make decisions and act in an environment to maximize a cumulative reward signal. In the context of autonomous agents, RL provides a way for these agents to learn how to perform tasks and make decisions without explicit programming by learning from trial and error. The autonomous agent is the learner in the RL framework. It interacts with an environment and takes action to achieve certain goals. The environment is the external system with which the agent interacts. It provides feedback to the agent through rewards based on the agent's actions. At each time step, the environment is in a certain state, which represents a snapshot of the environment's current condition. The agent selects an action based on its current state. The action influences the environment, causing it to transition to a new state and rewarding the agent. The environment provides a numerical reward signal to the agent after each action, indicating the immediate benefit or cost of the action. The agent's policy is a strategy that maps states to actions. The agent's goal is to learn an optimal policy that maximizes the cumulative reward over time. The value function estimates the expected cumulative reward that an agent can achieve from a given state while following a particular policy. It helps the agent make informed decisions about which actions to take. The agent uses an RL algorithm to learn the optimal policy or value function through trial and error. The learning process involves exploring the environment, receiving rewards, and adjusting the policy to improve performance. The agent faces a trade-off between exploring new actions to discover potentially better strategies (exploration) and exploiting known actions to maximize immediate rewards (exploitation). However, RL for autonomous agents also comes with challenges, such as efficient exploration, long-term rewards, and safe learning in real-world environments. \\
One of the main challenges in reinforcement learning is managing the trade-off between exploration and exploitation. To maximize the rewards, an agent must exploit its knowledge by selecting actions that are known to result in high rewards. On the other hand, it has to take the risk of trying new actions, which may lead to higher rewards than the current best-valued actions for each system state. In other words, the learning agent has to exploit what it already knows to obtain rewards, but it also has to explore the unknown to make better action selections in the future. RL uses strategies based on $\epsilon$-greedy and softmax to manage and tune this trade-off. Generally, the agent should explore more at the beginning of the training process when little is known about the problem environment. For RL agents, Markov Decision Processes (MDPs) are used to formalize sequential decision-making problems. An MDP consists of a set of $S$ states, a set of $A$ actions, a transition function $T$, and a reward function $R$, giving a tuple of $<S, A, T, R>$. RL-Based approaches can be divided into Value-based, policy-based, and actor-critic: \\

\subsubsection{Value Based RL}
Value-based Reinforcement Learning algorithms focus on estimating the value of different states or state-action pairs in an environment. The \emph{value} in this context represents how desirable or beneficial a particular state or action is for an agent with respect to achieving its long-term goals. The primary goal of value-based RL is to learn an optimal value function that guides the agent's decision-making process. The state-value function, denoted as $V_s$, estimates the expected cumulative reward an agent can obtain from a given state s while following a certain policy. It quantifies how good it is for the agent to be in a particular state. The action-value function, denoted as $Q(s, a)$, estimates an agent's expected cumulative reward from being in state s, taking action a, and following a certain policy. It quantifies the goodness of taking a specific action from a specific state. The Bellman equation describes the relationship between a state's value and its neighboring states' values. It is a fundamental equation used in value-based RL algorithms to update the value estimates based on expected future rewards. The optimal value function represents the maximum achievable value for each state, assuming the agent follows the best possible policy. The goal of value-based RL is to approximate this optimal value function. A common value-based RL algorithm is Q-learning \cite{q-learn}, an off-policy algorithm that iteratively updates action-value estimates based on the Bellman equation. It aims to find the optimal Q-function.
\begin{equation}
    q_\pi(S_t, a_t) \leftarrow q_\pi(S_t, a_t) + \alpha(Y-q_\pi(S_t, a_t))
\end{equation}
where Y is the temporal difference target and $\alpha$ is the learning rate. Deep Q-learning (DQN) is an extension of Q-learning that uses neural networks to approximate the Q-function. It is effective for handling high-dimensional state spaces. \cite{dql} from DeepMind uses this approach to learn to play Atari. The double-q-learning algorithm \cite{dqn} addresses the issue of overestimation of action values that can occur in standard Q-learning. The prioritized experience replay \cite{prioritized} is an enhancement to DQN that prioritizes the replay of experiences based on their importance for learning. \\

\subsubsection{Policy Based RL}
 Value-based RL revolves around approximating the value of different states or state-action pairs, indicating how good they are for an agent's goals. The agent subsequently deduces its actions from these values. Policy-based RL, in contrast, directly aims to learn the optimal decision-making policy that maximizes cumulative rewards by mapping states to actions. Value-based methods necessitate exploration mechanisms to discover actions and states, while policy-based methods inherently include exploration through stochastic policies. The stability and convergence of both approaches differ, with value-based methods facing challenges, especially when involving neural networks, and policy-based methods generally offering smoother updates. Policy-based methods directly search and optimize a parameterized policy $\pi_\theta$ to maximize the expected return:
 \begin{equation}
     max_\theta J(\theta) = max_\theta v_{\pi_\theta}()S = max_\theta \mathbb{E}_{\pi_\theta}[G_t|S_t=S]
 \end{equation}
where $\theta$ denotes the policy parameters. One problem of policy-based methods is poor gradient updates may result in newly updated policies that deviate wildly from previous policies, which may decrease the policy
performance. Trust region policy optimization (TRPO) \cite{trpo} prevents the updated policies from deviating too much from previous policies, thus reducing the chance of a bad update. TRPO optimizes a surrogate objective function where the basic idea is to limit each policy gradient update as measured by the Kullback-Leibler (KL) divergence between the current and the new proposed policy. This method results in monotonic improvements in policy performance. On the other hand, Proximal Policy Optimizations (PPO) \cite{ppo} clips surrogate objective function by having a too-large policy change. Inspiration is taken from gradient clipping here. However, it achieves a similar advantage in not deviating too much from the past time step. \\

\subsubsection{Actor-Critic RL}.
The Actor-Critic method is a hybrid reinforcement learning approach combining policy- and value-based elements. It aims to address the limitations of each approach by leveraging their strengths. In the Actor-Critic framework, an agent simultaneously learns a policy (actor) and an estimate of the value function (critic). In this approach, the actor (policy) selects actions based on the current state. It learns a policy that maps states to actions to maximize the expected cumulative reward. The policy can be stochastic, allowing for exploration. Common representations for the actor include neural networks. The critic (value Function) estimates the value of states or state-action pairs. It provides feedback to the actor by evaluating how good the current policy is. The value function helps guide the actor's learning process by providing a baseline for comparing different policies. The actor-critic method combines the benefits of policy-based methods (smooth learning and direct policy optimization) with value-based methods (efficient exploration and handling of continuous action spaces). An important concept in the actor-critic method is the advantage function, which represents how much better or worse an action is compared to the average action in a given state. The advantage function guides the actor's policy updates. Deterministic Policy Gradient Algorithms (DPG) \cite{dpg} is an off-policy actor-critic algorithm that derives deterministic policies. Unlike stochastic policies, DPG only integrates over the state space and requires fewer samples in problems with large action spaces. 
 
\subsection{Teacher Student Paradigm}
State-of-the-art methods for end-to-end autonomous driving follow the \emph{teacher-student} paradigm. The Teacher model, such as CARLA \cite{carla}, uses privileged information (ground-truth states of surrounding agents and map elements) to learn the driving strategy. The student model only has access to raw sensor data and conducts behavior cloning on the data collected by the teacher model. 

\begin{figure}[ht]
\vskip 0.2in
\begin{center}
\centerline{\includegraphics[width=\columnwidth]{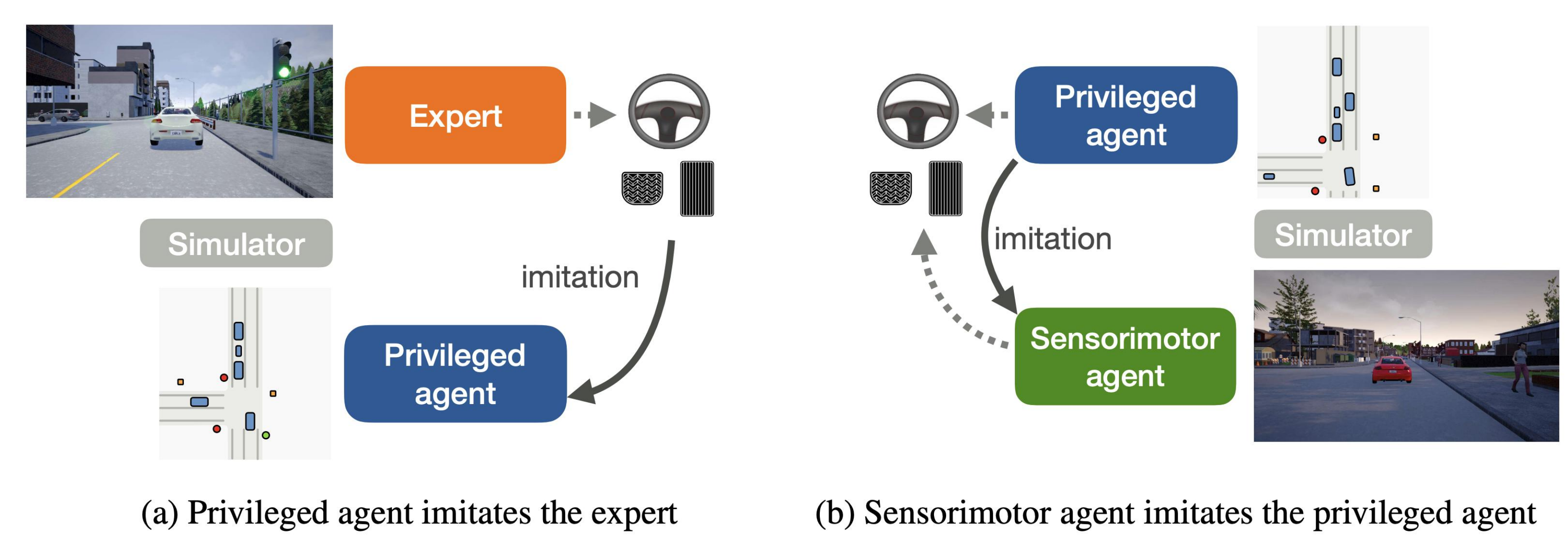}}
\caption{Demonstration of Teacher Student Paradigm from \cite{distillation_2}}
\label{fig:distillation}
\end{center}
\vskip -0.2in
\end{figure}

In \cite{distillation_2}, authors break up the complex task of perception and planning into two parts. In the first stage, they train an agent with access to privileged information. This privileged agent cheats by observing the environment's ground-truth layout and all traffic participants' positions. This agent learns a robust policy by cheating. It does not need to learn to see because it gets direct access to the environment’s state.  In the second stage, the privileged agent acts as a teacher who train a purely vision-based sensorimotor agent. The resulting sensorimotor agent does not have access to any privileged information and does not cheat. The privileged agent is a
“white box” and can provide high-capacity on-policy supervision. This architecture can be seen in Fig. \ref{fig:distillation}.

\cite{distillation_1} points out the gap under the current Teacher-Student paradigm, that the student model still needs to learn a planning head from scratch, which could be challenging due to the redundant and noisy nature of raw sensor inputs and the causal confusion issue of behavior cloning. Their approach explores the possibility of directly adopting the strong teacher model to conduct planning while letting the student model focus more on the perception part.\\
Though much effort has been devoted to designing a more robust expert and transmitting knowledge from teachers to students at various levels, the teacher-student paradigm still suffers from inefficient distillation.

\subsection{End to end Autonomous driving with the auxiliary task}
\begin{figure}[ht]
\vskip 0.2in
\begin{center}
\centerline{\includegraphics[width=\columnwidth]{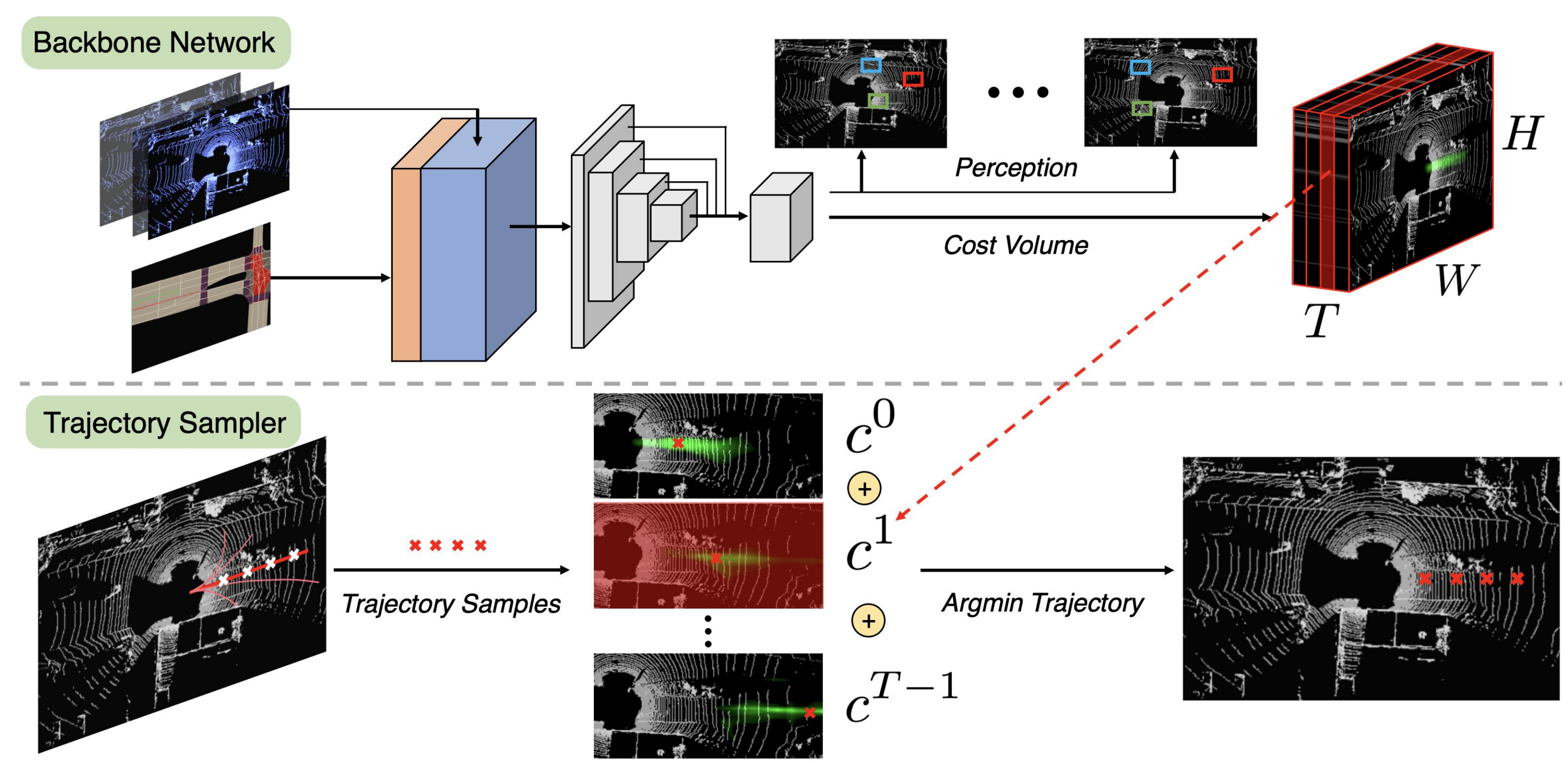}}
\caption{\cite{e2e_atg} End-to-end autonomous driving system with auxiliary training with perception losses}
\label{fig:auxilliary}
\end{center}
\vskip -0.2in
\end{figure}

Generally, it is commonly believed that learning control output/ waypoints for trajectory planning might be too complex to be mapped from the billions of parameters from the sensor data. This has led researchers to develop auxiliary tasks to supervise the learning end-to-end task as we have seen previously in \cite{distillation_1} where perception and planning modules are trained in the two stages. Another more discrete approach is covered in the paper by Uber ATG \cite{e2e_atg}, where they learn end-to-end interpretable motion planner using a backbone that takes LiDAR data and maps as inputs. First, they output bounding boxes of other actors for future timesteps (perception) and a cost volume for planning with T filters. Next, for each trajectory proposal from the sampler, its cost is indexed from different filters of the cost volume and summed together. The trajectory with the minimal cost will be our final planning as shown in Fig. \ref{fig:auxilliary}. They use a multi-task
training with supervision from detection, motion forecasting, and human-driven trajectories for the ego car. Note there is no supervision for the cost volume. Thus, they adopt max-margin loss to push the network to learn to discriminate between good and bad trajectories. The overall loss function is then:
\begin{equation}
    L = L_{perception} + \beta L_{planning}
\end{equation}
This multi-task loss directs the network to extract useful features, making the network output interpretable results. This is crucial for self-driving as it helps understand failure cases and improve the system. Here, perception loss is based on a classification loss, regression loss, and IoU (intersection over Union) loss for bounding boxes as shown in Fig. \ref{fig:auxilliary}. For Planning loss, they use the max-margin loss where we use the ground-truth trajectory as a positive example and randomly sampled trajectories as negative examples. The intuition behind this is to encourage the ground-truth trajectory to have minimal costs and others to have higher costs. \cite{e2d_wayve_2} is another method that uses a segmentation mask, monocular depth, and optical flow as an intermediate prediction for an end-to-end task of predicting control output using imitation learning.

\section{Open Challenges and Future Trends}
Considering challenges in the methods discussed and taking inspiration from the multi-disciplinary field of machine learning, we came across potential research directions for future researchers:
\begin{itemize}
    \item Collaborative end-to-end task: There has been research on collaborative perception, where perception is performed from the autonomous cars in the vicinity jointly to deal with the occlusions and extend the detection range. A similar concept is yet to be seen for end-to-end autonomous driving tasks. With more and more autonomous car companies deploying a fleet of cars in a city, this collaborative end-to-end task can be the next paradigm for SOTA methods in the field.
    \item Generate Scenes based on Text Requirements using Diffusion Models: Autonomous vehicles have been trained to work very well on generic scenarios; one of the major challenges still unaddressed is the long-tail problems, which we can't model in the simulators extensively. Recently, there have been great results with image-generation models using the stable-diffusion method \cite{stable-diffusion}; it would be great to see future researchers leverage this technology to build more realistic simulation scenarios. The end-to-end task of autonomous driving can be trained in an end-to-end task where diffusion models automatically generate similar scenes of the failure cases of autonomous vehicles. Then, we can train on those cherry-picked simulated scenes until our model predicts as expected on the real-world metrics.
    \item Foundation Models: Recently, foundation models in computer vision \cite{cv_foundation} and languages \cite{nlp_foundation} have shown state-of-the-art performance after being trained on the diverse, large-scale dataset and can be adapted to a wide range of tasks (driving scenarios). Researchers can leverage these models for generalizing capability for end-to-end tasks from the simulators to the real-life data. 
\end{itemize}

\section{Conclusion}
In conclusion, this review paper has comprehensively explored the rapidly evolving field of end-to-end autonomous driving. The journey from traditional modular approaches to the paradigm-shifting concept of end-to-end learning has been illuminated, showcasing the shift from handcrafted features and intricate pipelines to data-driven solutions that allow vehicles to learn directly from raw sensor inputs. The discussion spanned the advantages and challenges of this approach, highlighting its potential to simplify system design, capture complex interactions, and adapt to diverse driving scenarios. As this field continues to mature, the paper discussed the importance of safety assurances, regulatory frameworks, and the need for collaboration among academia, industry, and policymakers. Open challenges were identified, such as safety in extreme scenarios, transfer learning, interpretability, human interaction, and efficient exploration, inviting researchers and practitioners to collectively contribute to shaping the future of autonomous driving.

In essence, this review paper has showcased how end-to-end autonomous driving is more than just a technological leap; it is a paradigm shift that embodies the fusion of machine learning, robotics, and transportation engineering. The road ahead is paved with exciting opportunities and complex challenges, demanding an interdisciplinary effort to realize the vision of safe, efficient, and reliable autonomous vehicles that seamlessly navigate our modern roadways.

\addtolength{\textheight}{-12cm}   




{\small
\bibliographystyle{ieee_fullname}
\bibliography{IEEEexample}
}

\end{document}